\documentclass[10pt, a4paper]{article}
\usepackage{lrec-coling2024} 

\usepackage{url}
\usepackage{amsmath}
\usepackage{graphicx}
\usepackage{xcolor}
\usepackage[capitalize]{cleveref}
\usepackage{tabularx}
\usepackage{booktabs}
\usepackage{xspace}
\usepackage{floatrow}
\usepackage{enumitem}
\usepackage[font=small]{caption}

\setlist[enumerate]{itemsep=0mm}

\graphicspath{{images/}}

\renewcommand{\UrlFont}{\ttfamily\small}

\makeatletter
\newcommand\footnoteref[1]{\protected@xdef\@thefnmark{\ref{#1}}\@footnotemark}
\makeatother

\usepackage{xspace}

\name{Oana Ignat, Longju Bai, Joan Nwatu, Rada Mihalcea}

\address{University of Michigan \\ 
\tt \{oignat,longju,jnwatu,mihalcea\}@umich.edu}

\title{Annotations on a Budget: Leveraging Geo-Data Similarity to \\
Balance Model Performance and Annotation Cost}

\abstract{
Current foundation models have shown impressive performance across various tasks. However, several studies have revealed that these models are not effective for everyone due to the imbalanced geographical and economic representation of the data used in the training process. Most of this data comes from Western countries, leading to poor results for underrepresented countries. To address this issue, more data needs to be collected from these countries, but the cost of annotation can be a significant bottleneck. 
In this paper, we propose methods to identify the data to be annotated to balance model performance and annotation costs. Our approach first involves finding the countries with images of topics (objects and actions) most visually distinct from those already in the training datasets used by current large vision-language foundation models. Next, we identify countries with higher visual similarity for these topics and show that using data from these countries to supplement the training data improves model performance and reduces annotation costs. The resulting lists of countries and corresponding topics are made available at {\UrlFont \url{https://github.com/MichiganNLP/visual_diversity_budget}}.
 \\ \newline }

\begin{document}

\maketitleabstract

\section{Introduction}
Vision-language models have shown remarkable advances in recent years~\citep{Li2019VisualBERTAS, Zhang2021VinVLRV, Radford2021LearningTV, Zellers2021MERLOTMN, Li2022BLIPBL, Kirillov2023SegmentA, huang2023tag2text}. These models have shown great performance on a variety of tasks, from lower-level tasks such as object detection, image segmentation~\citep{Kirillov2023SegmentA}, and image and video classification to higher-level tasks such as
image/ video captioning~\citep{Li2022BLIPBL, huang2023tag2text}, text-image/video retrieval~\citep{Radford2021LearningTV}, visual question answering and visual commonsense reasoning~\citep{Zellers2021MERLOTMN, Zellers2022MERLOTRN}.

At the same time, prior work has demonstrated that these models do not work well for everyone~\citep{Devries2019DoesOR}. Specifically, models do not work well on out-of-domain data, and data from low-income and non-western countries~\cite{nwatu2023bridging}.
This is due to the imbalanced geographical and economic representation of the data used to train these models, as it comes mainly from North America and Western Europe \citep{Shankar2017NoCW}.
One solution that \citet{Rojas2022TheDS} and \citet{Ramaswamy2023BeyondWC} propose is to collect more data from underrepresented countries. However, as \citet{Ramaswamy2023BeyondWC} highlights, annotation costs are a substantial bottleneck; when crowdsourcing the data, fair pay is about 1.08\$ per image without including researcher time.

\begin{figure}[h!]
    \centering
    \includegraphics[width=\textwidth]{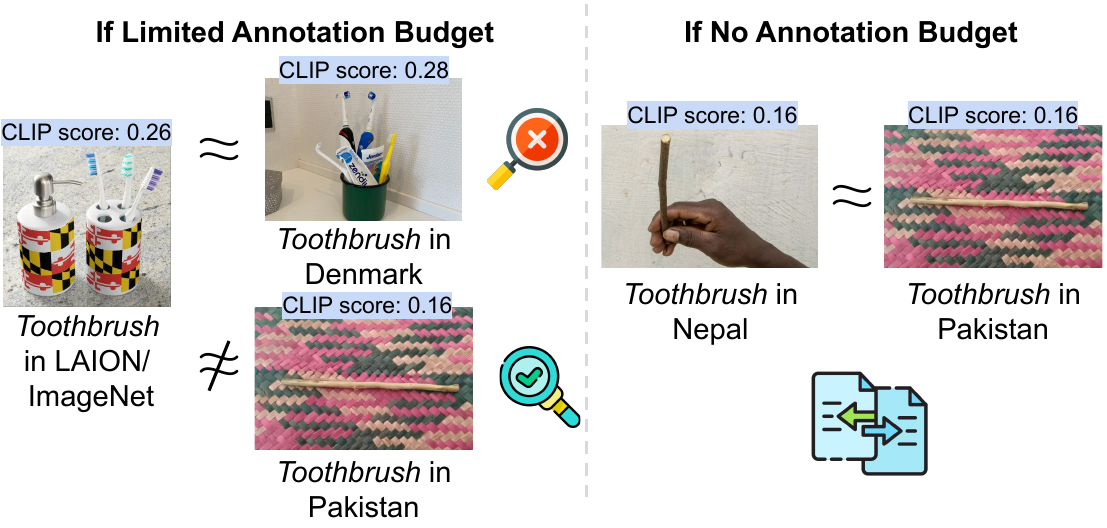}
    \caption{Vision-language models work poorly on data from underrepresented countries. This is primarily due to the diverse appearance of topics (objects and actions) across countries (e.g., ``toothbrush''). 
    However, collecting diverse global data is very expensive. As solutions to budget annotations, we propose to (1) annotate the images visually different from the ones in high-resource datasets such as LAION or ImageNet; (2) supplement data from low-resource countries with data from visually similar countries.} 
    \label{fig:income_boxplot}
\end{figure}

As a complementary solution, we investigate strategies to reduce the annotation budget while finding effective annotation data.  
Specifically, our paper aims to answer two main research questions.

\begin{description}[leftmargin=*]
\item [RQ1:] {\bf Which countries are less represented in the training data of vision-language models?}

We aim to find ways to effectively focus future annotation efforts on specific \textit{countries} and their corresponding \textit{topics} (objects and actions).\footnote{Throughout the paper, for brevity, we use the term \textit{country} to refer to a country or territory.} 
Our study highlights the visual diversity of common topics across countries and those that differ the most from the primarily Western data used to train most multimodal foundation models.

\item [RQ2:] {\bf How can we leverage cross-country data similarity to improve the representation of vision-language models?}

We obtain groups of countries that are visually similar in their representation of a given topic. This is particularly useful when there is not enough data for one of the countries in the group, and there is no annotation budget. We can supplement the data from this country by using data from the other countries in the group. 
\end{description}

\noindent We summarize our contributions as follows. First, \textbf{we identify the data likely to most benefit from annotations} by finding which countries and corresponding topics are less represented in the training data of vision-language models.
Second, across 52 countries and 94 topics, \textbf{we identify the groups of countries that are visually similar in their representation of a topic and show that they can be used to supplement training data effectively}. 
Third, \textbf{our main takeaways create opportunities for affordable and geo-diverse data collection}, encouraging contributions to creating datasets and models that work for everyone.

\section{Related Work}


\paragraph{Data Subset Selection/ Active Learning}
There have been numerous studies on the use of semi-supervised models to leverage a combination of limited labeled data and vast amounts of unlabeled data to improve model performance at lower costs \citep{hady2013semi, oliver2018realistic, taha2023semi, chen2022semi}. 

However, model-generated labels could be inconsistent and unrepresentative with semi-supervision, leading to reduced model performance \citep{ahfock2023semi, elezi2022not, Wang2021UnsupervisedSL}. While similar to semi-supervised learning in objective, active learning methods seek to capture the entire data distribution by focusing labeling efforts on the data points that provide the most information for training the best-performing models \citep{ren2021survey, citovsky2021batch, monarch2021human, yang2017suggestive} using approaches such as uncertainty-based sampling in \citet{gal2016dropout, beluch2018power}  and geometric-based methods in \citet{sener2018active}. Unsupervised subset selection methods like K-means and K-median core set in \citet{har2005smaller}, which form the foundation for geometric-based active learning approaches are similar to our work which seeks to select a subset that is representative of the entire dataset using distance metrics. However, the objective of the selection is to include images from a low-resource dataset with the least similarity to data of the same class in a high-resource dataset.

\paragraph{Evaluating Disparities in Model Performance}
There exists a considerable body of literature evaluating the fairness and the unequal performance of vision and vision-language models on diverse groups categorized according to race \citep{gebru2020race}, gender \citep{Buolamwini2018GenderSI}, geolocation \citep{Kim2021TowardsAF, Shankar2017NoCW, Goyal2022VisionMA} and income \citep{Devries2019DoesOR, nwatu2023bridging}.  

Further analysis of these disparities reveals that factors such as ambiguous label definitions, domain shifts, annotator disagreement \citep{hall2023towards, kalluri2023geonet}, as well as image properties relating to texture, lighting, and occlusion in vision and vision-language datasets \citep{Gustafson2023PinpointingWO} contribute to disparities in datasets which carry over to affect model performance. 

Frameworks have been developed to facilitate the detection of bias through guided human-in-the-loop inspection, either in datasets \citet{HuCrowdsourcing} or in models \citet{Goyal2022FairnessIF}.  
Our work focuses on exploring the presence of variations in image representations across demographic groups in existing datasets, to inform cost-effective methods for building balanced, diverse datasets.
\paragraph{Improving Representation in AI. }
Efforts toward improving equal representation in AI and equitable AI impact revolve around model adaptation, transfer learning, and dataset diversity. However, \citet{salman2022does, kalluri2023geonet, Dubey2021AdaptiveMF, wang2023overcoming} suggest that transfer learning and model adaptation methods might not be enough to eradicate the issue of under-representation in AI models. 

On the other hand, adding diverse data to training datasets tends to yield significant improvements in model performance across different groups \citep{Ramaswamy2023BeyondWC, Rojas2022TheDS}. The need for more diverse datasets has become apparent, leading to the development of datasets like GeoYFCC \citep{Dubey2021AdaptiveMF}, GeoDE \citep{Ramaswamy2023BeyondWC}, Dollar Street \citep{Rojas2022TheDS}, and Segment Anything \citep{kirillov2023segment} that include data collected from diverse locations. 

While advantageous, diverse datasets are expensive and resource-intensive to build. \citep{schumann2021step,garcia2023uncurated, geigle2023babel} explored a less expensive alternative: revising or creating annotations for an existing dataset to improve inclusivity and reduce bias.  
Similarly, we seek to facilitate effective but less expensive annotations by leveraging the differences between high-resource and low-resource datasets to curate the best low-resource subset for annotation.

\section{Methodology}
We start by collecting two datasets that reflect the low-resource and high-resource settings. First, we compile a crowd-sourced geo-diverse dataset collected from a large number of countries, which we refer to as ``low-resource data'' due to the low number of images that could be collected for each country in the set and the difficulty of gathering more. Second, we also compile a web-scraped dataset used for training foundation models, which we refer to as ``high-resource'' due to its vast size consisting of billions of images (e.g., LAION-5B\footnote{\url{https://laion.ai/blog/laion-5b/}}) and the ease of gathering more data.

Next, we pre-process the data by mapping the topics between the two data sources, filtering out topics and countries with very few images.
Finally, we utilize the collected data to generate visual representations through vision-language foundation models. These representations are then used to determine the visual similarity between images of topics in low-resource data and their corresponding topics in high-resource data.

\subsection{Low-resource Multimodal Data}

We combine two geographically diverse datasets: GeoDE~\citep{Ramaswamy2023BeyondWC} and Dollar Street~\citep{Rojas2022TheDS}.
For brevity, we call \textit{topics} all the labels used for all the objects and actions in these two datasets.

\noindent{\bf GeoDE.} 
The GeoDE dataset contains $61,940$ crowd-sourced images of $40$ objects. The data is balanced across six regions (West Asia, Africa, East Asia, South East Asia, Americas, and Europe), each with 3-4 countries. These regions were chosen due to their scarcity in most public datasets. 
Using a combination of heuristics and manual validation, the authors selected the objects likely to be visually distinct across the six regions.

\noindent{\bf Dollar Street.} 
The Dollar Street dataset contains $38,479$ images collected from $63$ countries on four continents (Africa, America, Asia, and Europe).  The images capture everyday household objects and actions (e.g., ``toothbrush'', ``toilet paper'', ``cooking'').
The data contains $291$ unique topics, out of which we remove nineteen subjective topics following the work of \citet{Devries2019DoesOR} (e.g., ``most loved item'', ``things I wish I had''). 
All the subjective topics are found in the Appendix. 
The number of images for a given country ranges from $45$ in Canada to $4,704$ in India, with a median of $407$ images per country.

\subsection{High-resource Multimodal Data}

As high-resource datasets, we sample data from ImageNet~\citep{Deng2009ImageNetAL} and LAION~\citep{Schuhmann2022LAION5BAO}.
We chose these datasets due to their popularity in vision-language models.
\paragraph{ImageNet.} ImageNet and ImageNet Large Scale Visual Recognition Challenge (ILSVRC) are pioneers in advancing object detection and classification progress.
The imagenet21k dataset~\citep{Deng2009ImageNetAL} contains around 21,000 WordNet~\citep{Fellbaum2000WordNetA} synsets and more than 14 million annotated images. We use the processed version of ImageNet21k~\citep{Ridnik2021ImageNet21KPF}, with removed invalid classes and resized images. We also tried using ImageNet1k, but it did not have enough classes for our purpose, and we chose to use it to supplement the ImageNet21k data.
\paragraph{LAION.} Large language-vision models such as CLIP or ALIGN
have been trained on billions of image-text pairs unavailable to the public. LAION-5B~\citep{Schuhmann2022LAION5BAO} was created to address this problem by open-sourcing a CLIP-filtered dataset\footnote{The data is filtered using OpenAI's CLIP ViT-L/14 by calculating the cosine similarity between the text and image embeddings and dropping those with a similarity below 0.3.
} of 5,85 billion high-quality image-text pairs.  
We use LAION-400M~\citep{Schuhmann2021LAION400MOD}, a subset of LAION-5B that contains 400 million English image and text pairs.

\subsection{Data Pre-processing}

\paragraph{Combine GeoDE and Dollar Street.} 
We pre-process and combine the low-resource datasets to increase the number of topics, images, and country diversity.
First, we manually group and rename the topics from Dollar Street with the same meaning (e.g., ``bathroom privacy'', ``bathroom/ toilet'' are renamed ``bathroom'').
Next, we rename the topics from Dollar Street that match those in GeoDE (e.g., ``bike'' to ``bicycle'', ``medication'' to ``medicine'').
We remove three topics with less than $10$ images per topic.  
Finally, we obtain a total of $99$ unique topics, $93,060$ images, from $4$ continents, $18$ regions, and $83$ countries. 
\paragraph{Low-resource to High-resource Data Mapping.}

We map the $99$ topics from the low-resource data to the high-resource data, ImageNet, and LAION by identifying the images with similar labels.

First, we map $51$ topics from the low-resource data to an exact match to ImageNet21k or ImageNet1k. We could not find an exact match for $38$ topics because these topics are too abstract (e.g., ``jewelry'', ``source of cool'', ``religious building''). Instead, we find mappings for their hyponyms (e.g., for ``jewelry'', we map ``bangle'', ``necklace'', ``bracelet'' and ``ring''). The remaining $10$ topics for which we could not find any exact or hyponym mapping to ImageNet21k or ImageNet1k are mapped to LAION.

We map data in LAION by selecting the images with captions that contain the topic query. Because LAION data is web-crawled, we find that the images are lower quality than ImageNet and not always relevant to the topic query: e.g., the ``TV'' topic in LAION contains images of people on TV, not of the object TV. Therefore, to ensure the correctness of the mapping, we manually inspect the images and map a topic to LAION only when most images are relevant to the topic query. We map $64$ topics to LAION.
Note, however, that the number of hyponyms and the quality of LAION images limit how comprehensive the mapping process is. 
Two independent annotators check 20 random images from each topic and find that most noisy images come from LAION. Therefore, we decide to limit the amount of data from LAION and add more images from ImageNet. 
Specifically, we randomly sample around $200$ images per topic from LAION and around $1,000$ images per topic from ImageNet. Note that the high-resource data does not contain country information.
We show the data before and after pre-processing and the topic mapping in our repository.\footnote{\label{note1}\UrlFont \url{https://github.com/MichiganNLP/visual_diversity_budget}}

\begin{figure}[h]
    \centering
    \includegraphics[width=\textwidth]{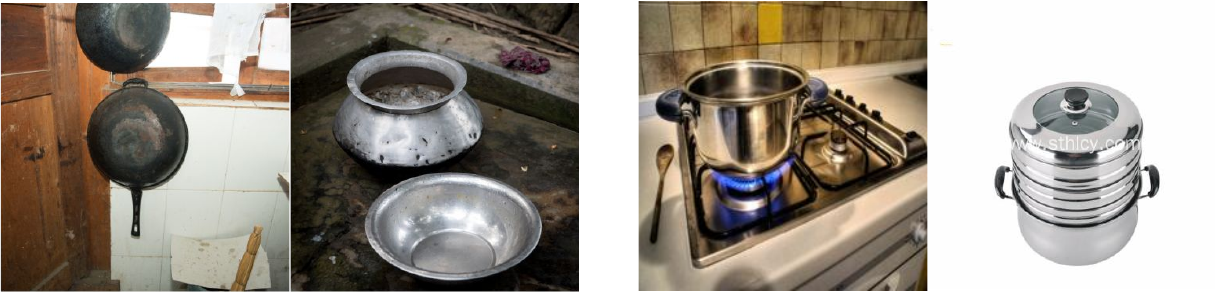}
    \caption{Example images (``cooking pot'') in low-resource data (left) vs. in high-resource data (right).}
    \label{fig:low_vs_high}
\end{figure}

\paragraph{Data Filtering.} 
The low-resource data is unbalanced, as the total number of images per country varies from 6,549 for Japan to 1 for Bulgaria and Venezuela, with a median of 345 images per country. 
The number of images per topic is also unbalanced, from 3,049 for ``waste container'' to 18 for ``hanging clothes to dry''.
However, balancing the data by down-sampling significantly reduces the number of countries represented for each topic. Having numerous countries represented is essential for our setup. Therefore, we choose not to balance the data. 
Instead, we remove the (topic, country) pairs containing less than $10$ images, considering this threshold a minimum for experiment significance. This also removes considerable data: 3,329/ 4,830 (topic, country) tuple pairs, 5/ 99 topics, and 31/ 83 countries.
We show the removed topics and corresponding countries in our repository and 
highlight the need for more data for these pairs to obtain significant results.\footnoteref{note1}

We show the statistics after the data collection and pre-processing in \Cref{tab:data_stats} and the image distribution of countries per topic in Appendix \Cref{fig:topic_distrib}.

\begin{table}
    \small	
    \centering
    \begin{tabular}{lc}
    \toprule
    \# unique topics & 94 \\
    \# unique countries & 52 \\
    \# unique (topic, country) pairs & 1,501 \\
    \# images in low-resource data & 80,801 \\
    \# images in high-resource data & 103,006 \\
    average \# images per (topic, country) & 53.8 \\
    median \# images per (topic, country) & 30 \\
    \bottomrule 
    \end{tabular}
    \caption{Statistics for the collected number of topics, countries, and images collected from low-resource and high-resource data after data pre-processing.}
    \label{tab:data_stats}
\end{table}

\subsection{Data Representation}
We use an ensemble of three representations to compute the image similarity and to ensure the results generalize across representation types.
We choose CLIP~\citep{Radford2021LearningTV}, ALIGN~\citep{jia2021scaling}, and BLIP-2~\citep{li2023blip2} due to their popularity as foundation models~\citep{bommasani2022opportunities}, i.e., their use in a multitude of models and their high zero-shot performance across various tasks and datasets, such as text-to-image retrieval, image question answering, human action segmentation, image-sentence alignment, image captioning~\citep{Cafagna2021WhatVM,saharia2022photorealistic,kirillov2023segment, huang2023chatgpt}.

\paragraph{CLIP Representations.} 
We use the pre-trained Vision Transformer ViT-B/32~\citep{dosovitskiy2020vit} from the CLIP model~\citep{Radford2021LearningTV} to encode the visual representations of the images. The training dataset for CLIP was created from the results of numerous queries to various publicly available Internet sources. The dataset referred to as WebImageText WIT contains 400 million (image, text) pairs and is not available to the public. 

\paragraph{ALIGN Representations.} 
We also extract image features following the ALIGN \citep{jia2021scaling} model setup, using a pre-trained EfficientNet \citep{tan2019efficientnet} as a vision encoder.
Since the original code has not been released, our implementation is based on the Kakao Brain code that reproduced the original paper.\footnote{\small \url{https://huggingface.co/docs/transformers/model_doc/align}}
ALIGN was trained on 1.8 billion image-text pairs collected following the methodology used for the Conceptual
Captions dataset~\citep{sharma2018conceptual}. Since the emphasis was on scale instead of quality, the dataset underwent fewer post-processing steps, thus leading to a noisier dataset. This dataset is currently unavailable for public access. 

\paragraph{BLIP-2 Representations.} 
We also extract image features using BLIP-2~\citep{li2023blip2}, which uses ViT-g/14 from EVA-CLIP~\citep{sun2023evaclip} as image encoder and removes the second last layer's output features to increase the performance. BLIP-2 was trained on a total of 129M images aggregated from the COCO~\citep{lin2014microsoft}, Visual Genome~\citep{krishna2017visual}, CC3M~\citep{sharma2018conceptual}, CC12M ~\citep{changpinyo2021conceptual}, SBU~\citep{ordonez2011im2text}, and the LAION400M datasets~\citep{Schuhmann2021LAION400MOD}. Captions for the web images were generated using CapFilt~\citep{Li2022BLIPBL}. 

\section{Mapping the Representation of Vision-Language Models}

In this section, we address the first research question: {\bf 
RQ1: Which countries are less represented in the training data of vision-language models?}

For each (topic, country) pair, we compute the cosine similarity between 
the average visual representations of all the corresponding images in the low-resource data and the average visual representations of all the corresponding images in the high-resource data. Note that the average is computed over all three visual representation types, i.e., CLIP, BLIP, and ALIGN.
We select the (topic, country) pairs with a similarity score lower than a threshold computed as the average similarity score between all the image representations in the low-resource data and the corresponding representations in the high-resource data. This process is repeated for each visual representation type.\footnote{Thresholds and data representations can be changed to fit the purpose of the analysis or application.} 
Finally, the (topic, country) pairs selected for all three visual representations are the ones we find to be consistently different from the high-resource data and, thus, the ones that benefit the most from annotations.
We find 422 such (topic, country) pairs out of 1,501 unique (topic, country) pairs, potentially reducing the annotation budget to less than a third of the initial amount. We share the results in our repository.\footnoteref{note1}

\begin{figure*}[h!]
  \centering
  \includegraphics[width=\textwidth]{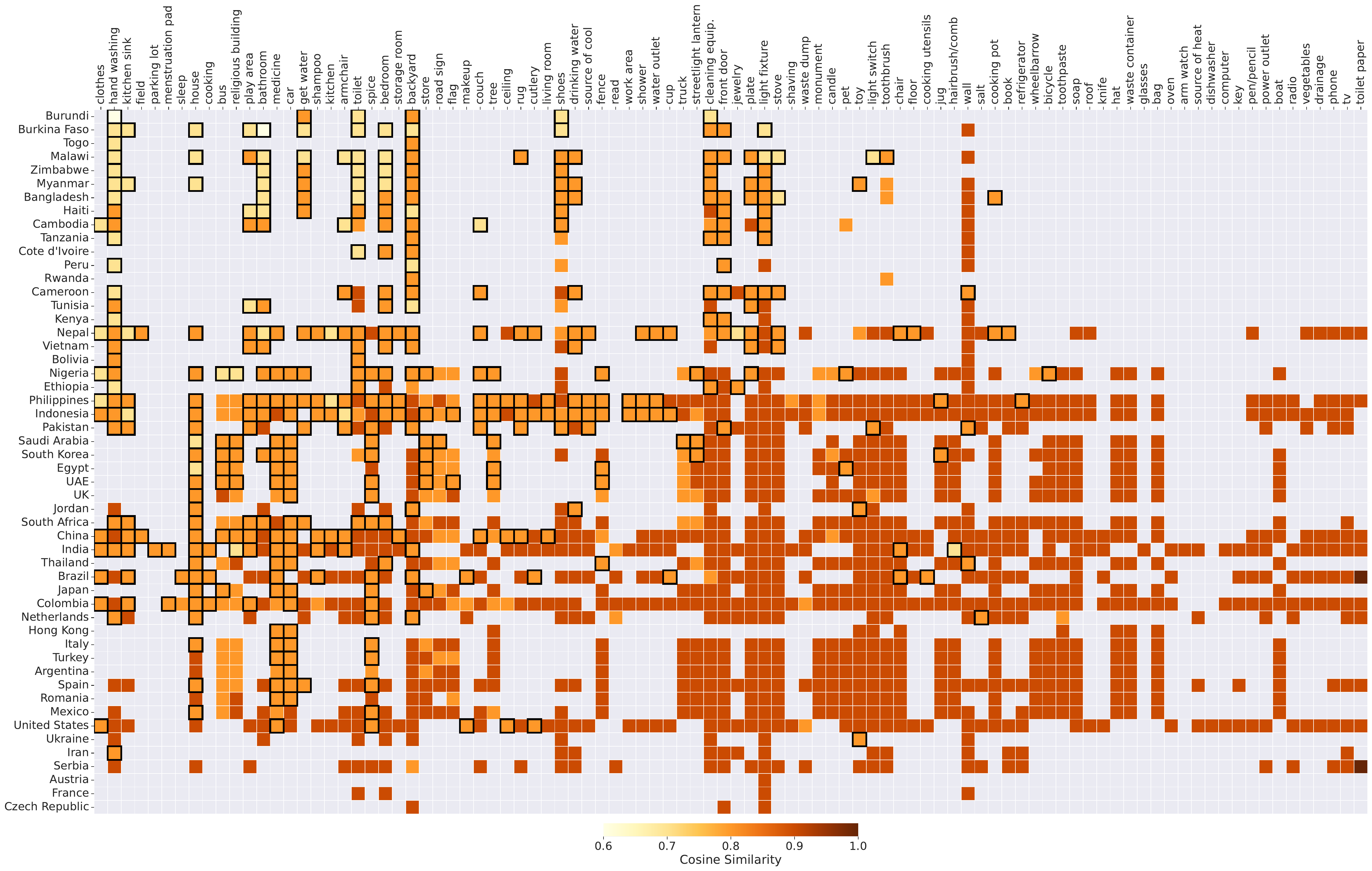}
  \caption{Similarity heatmap of (topic, country) pairs. Based on the average similarity score, rows and columns are sorted from the least to the most similar. The lighter the color, the lower the similarity between high-resource and low-resource data for that corresponding (topic, country) pair, the more beneficial it is to annotate. We highlight with \textit{black} the pairs we determine to benefit the most from annotations. Grey cells have less than ten images and are therefore discarded. \textit{Best viewed in color.}}
  \label{fig:CLIP_avg_results_highlight}
\end{figure*}

\paragraph{Visual similarity for each (topic, country) in low-resource data with corresponding topics in high-resource data.}
We compute a similarity heatmap where the rows are the topics, and the columns are countries.
We sort the rows (countries) and columns (topics) from the least to the most similar based on the average similarity score per country and topic, leaving out the $NaN$ values (the grey, empty cells).
We show in \Cref{fig:CLIP_avg_results_highlight} the similarity heatmap for the CLIP representation and highlight the (topic, country) pairs we find to benefit the most from annotations based on consistently low similarity with the high-resource data across the three visual representations. 

From \Cref{fig:CLIP_avg_results_highlight}, we can also see that the countries with the fewest data are usually the ones with the most topics in need of annotations (e.g., from $Burundi$ to $Kenya$). Exceptions to this are countries such as $Nepal$, $Nigeria$, $Philippines$, and $Indonesia$, which have more data points (topics), but more than half of the topics require annotations, and countries such as $Czech$ $Republic$, $France$ or $Austria$ which have very few topics and none require annotations.
In \Cref{fig:CLIP_avg_results_highlight}, we see a few topics in $United$ $States$ that are marked to require annotations: ``medicine'', ``spice'', ``ceiling'', ``clothes'' and ``makeup''. We show in Appendix \Cref{fig:usa_high_resource} representative images from these topics from the two data sources, which explain the visual differences.
For the rest of the topics, as expected, $United$ $States$ data is similar to the high-resource data. 
We considered using the $United$ $States$ as the high-resource data source. However, due to the lack of data on some topics and relatively few images per topic compared to other countries, it was not feasible.

There are differences between the results obtained with each visual representation type regarding similarity score intervals and which (topic, country) pairs are similar to the high-resource data. However, the general similarity trend is consistent as most (topic, country) pairs have only low or high similarity scores across all three representations. This is also supported by the strong Pearson correlations between the scores obtained with the three representation types: CLIP and BLIP scores correlate $0.62$, CLIP and ALIGN scores correlate $0.65$, ALIGN and BLIP scores correlate $0.72$. 
We show in the Appendix \Cref{fig:CLIP_avg_results}, \ref{fig:ALIGN_avg_results}, and \ref{fig:BLIP_avg_results}, the similarity heatmaps for each representation type: CLIP, ALIGN, and BLIP respectively. 

\begin{figure}[h]
    \centering
    \includegraphics[width=\textwidth]{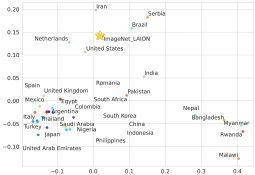}
    \caption{PCA for the topic ``toothbrush'' for all countries that contain this topic in the low-resource data and in the high-resource data. The high-resource data point is highlighted with \textit{star} symbol. The data is represented as the average of the CLIP representations. 
    }
    \label{fig:pca-rq1}
\end{figure}

\paragraph{Topic visual representation in high-resource and low-resource data.}
To show how the topic visual representations vary per low-resource and high-resource data,
we perform a 2D transformation using Principal Component Analysis (PCA)~\citep{doi:10.1080/14786440109462720}. 
In \Cref{fig:pca-rq1}, we show the CLIP average representations per country in the low-resource and the corresponding high-resource data for the topic ``toothbrush''.
We can observe that, for this topic, there is considerable visual diversity across countries. When comparing to the high-resource data, $ImageNet\_LAION$, we observe visually different countries, such as $Malawi$, $Rwanda$, and $Myanmar$, and countries very visually similar, such as $Netherlands$, $United States$, and $Brazil$.
In addition, we observe many countries that tend to be clustered together, i.e., visually similar for this particular topic, such as $Mexico$, $Italy$, $Japan$, $South$ $Korea$, and others.
We examine more about the similarities between countries when answering \textbf{RQ2}, in the following section.
In Appendix \Cref{fig:pca_hand-washing}, \ref{fig:pca_toilet}, \ref{fig:pca_wall} we show results for other topics (``hand washing'', ``toilet'', ``wall'') in low-resource and high-resource data.

\section{Cross-country Data Similarity for Improved Model Representation}

We now turn to the second research question {\bf RQ2: How can we leverage cross-country data similarity to improve the representation of vision-language models?}

We calculate the cosine similarity between the average visual representations of images for each topic across countries, and repeat this process for all three visual representations.
Given a topic, the final visual similarity score between two countries is obtained by averaging the similarity values obtained for each visual representation type. 
For each (country, topic) pair, we obtain the visually similar countries, along with their similarity score, from the most to the least similar, and share them in our repository.\footnoteref{note1}

We calculate the average similarity score for each country across all corresponding topics and for each topic across all corresponding countries.
We show the similarity score distribution for the top three and last three countries and topics in \Cref{fig:subset_sim},
and for all countries and topics in the Appendix \Cref{fig:scores_topic} and \ref{fig:scores_country}.

As shown in \Cref{fig:subset_sim}, $Burundi$ has the lowest similarity score of $0.775$, indicating that it is the most different country compared to the others and needs its own annotations.
On the other hand, $Argentina$ has the highest similarity scores of $0.907$, indicating a high similarity to other countries. These results imply that annotating data from $Argentina$ would help other countries.
The most visually different topic is ``religious building'' with a score of $0.76$, and the most similar topic is ``hat'' with a score of $0.96$. These results imply that ``religious buildings'' should be annotated more widely as their visual appearance varies across countries.

Finally, we investigate whether performance of similarity calculation depends on amount of annotated data. We find that at  topic level the similarity scores are not correlated with the amount of annotated data (Pearson correlation coefficient is -0.02). We discuss more about the effect of data size on our analysis results in the Appendix.

\begin{figure}[h]
    \centering
    \includegraphics[width=\textwidth]{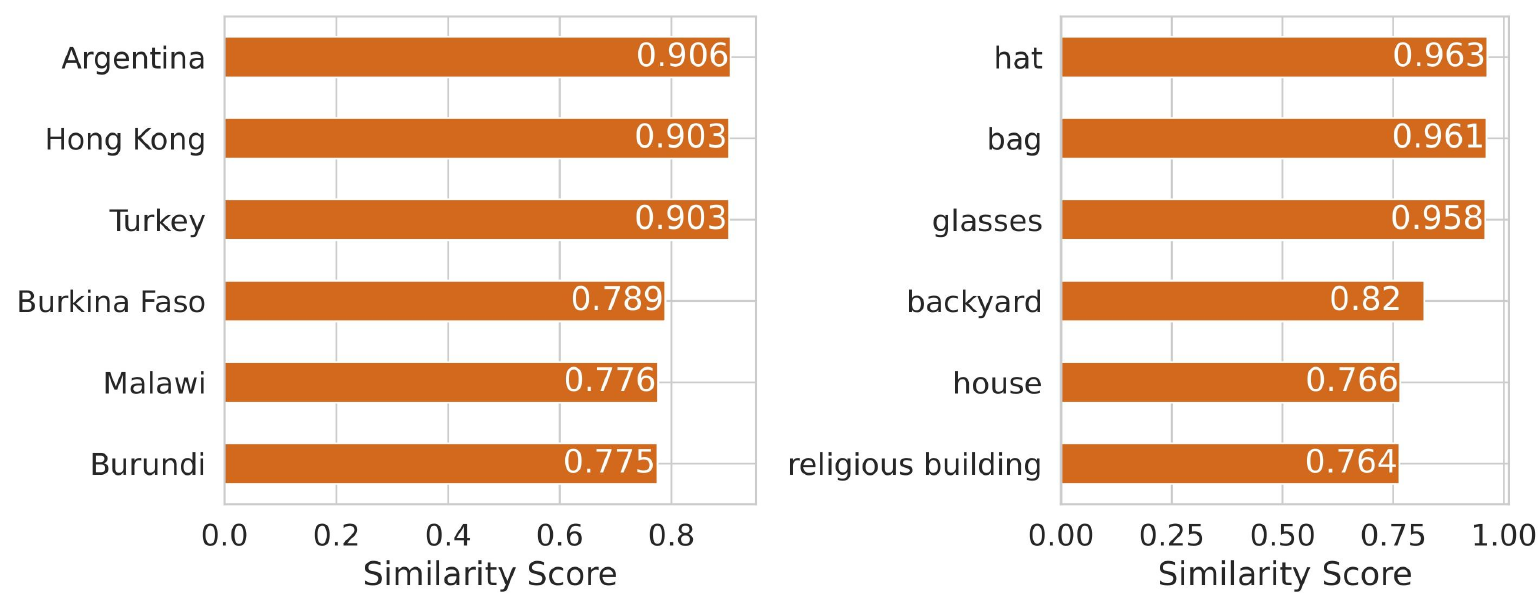}
    \caption{Top three and last three countries (left) and topics (right) sorted by average similarity score.
    }
    \label{fig:subset_sim}
\end{figure}

\begin{figure}[h]
    \centering
    \includegraphics[width=\textwidth]{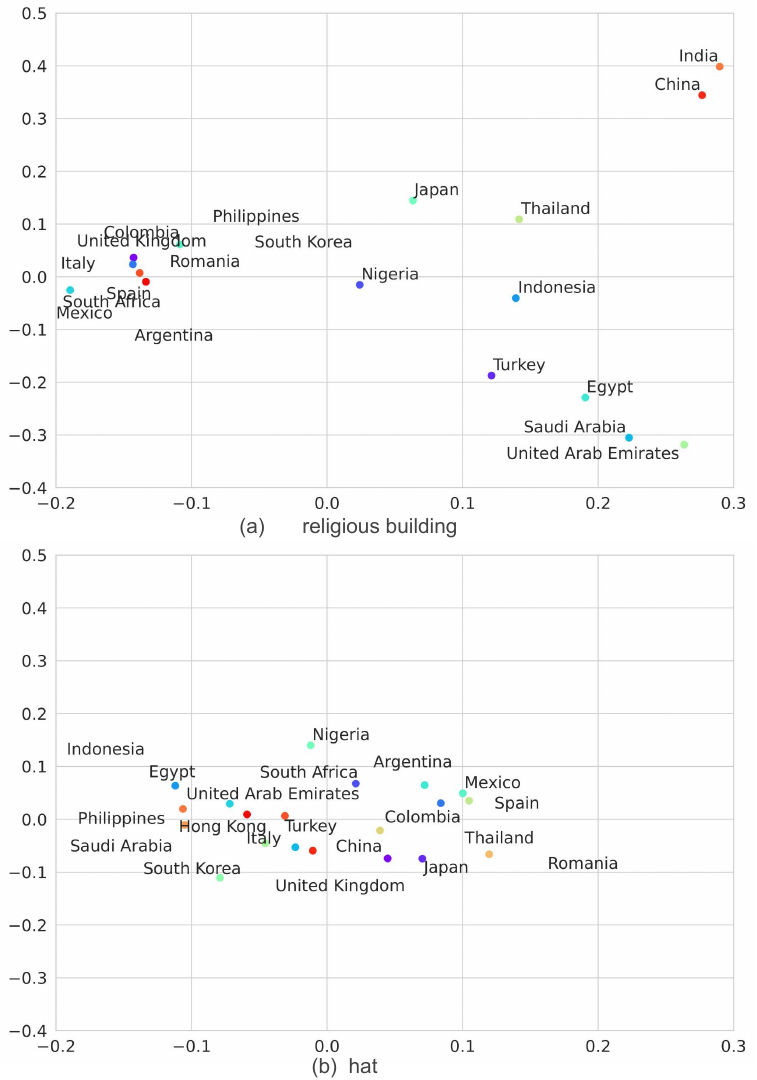}
    \caption{PCA for the topic ``religious building'' and ``hat'' for all countries in the low-resource data that contain this topic. The data is represented as the average of the CLIP representations.}
    \label{fig:pca-rq2}
\end{figure}

\paragraph{Topic visual representation across countries in low-resource data.}
To show how the topic visual representations vary per country in the low-resource data, we perform a 2D transformation using Principal Component Analysis (PCA)~\citep{doi:10.1080/14786440109462720}. 
In \Cref{fig:pca-rq2}, we show the CLIP average representations per country for the topics with the most and least visual differences across countries: ``religious building'' and ``hat'', respectively. As expected, the representations for ``religious building'' are much more spread across countries than those for ``hat'', which tend to cluster together. 
In Appendix \Cref{fig:get-water_PCA}, \ref{fig:house_PCA}, and \ref{fig:backyard_PCA}, we show representations for other topics visually different across countries: ``get water'', ``house'' and ``backyard.''

\paragraph{Correlation between geographical distance and visual similarity across countries.}
We measure if the visual similarity between countries correlates with the geographical distance. The geographical distance between two countries is calculated using Vincenty's distance~\cite{Vincenty1975DIRECTAI} between their capital cities.\footnote{\url{https://github.com/rahulbot/distances-between-countries}}
The visual similarity between any two countries is calculated across all their shared topics.
We compute the Pearson correlation coefficient~\cite{freedman2007statistics} over all countries and obtain a value of $-0.01$, indicating a weak negative correlation. A strong negative correlation is initially more expected as, intuitively, their visual similarity should increase as the distance between countries decreases. 
However, when we break down the correlation at the country level, the correlation coefficient varies significantly per country.
\begin{figure}[h!]
    \centering
    \includegraphics[width=\textwidth]{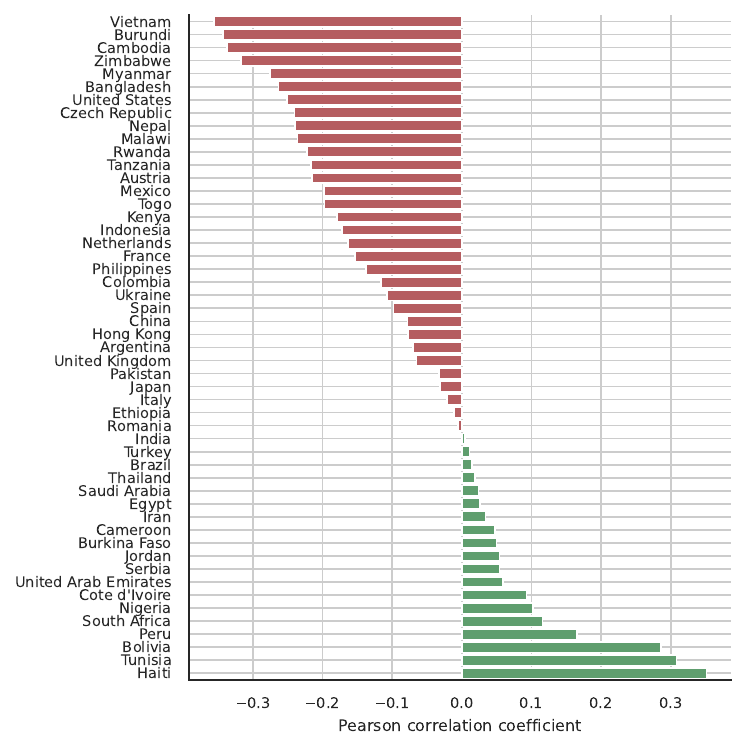}
    \caption{
    Pearson correlation coefficient between the visual similarity and the geographical distance, across countries. Most countries do not have a significant correlation between visual similarity and location.}
    \label{fig:distance_correlation}
\end{figure}
In 
\Cref{fig:distance_correlation}, we show countries with weak to moderate positive correlations (e.g., $Haiti$ with $0.35$, $Tunisia$ with $0.30$), countries with weak to moderate negative correlations (e.g., $Vietnam$ with $-0.35$, $Burundi$ with $-0.34$), most countries have values close to $0$, indicating no correlation between visual similarity and geographical distance.
Upon close examination of the results, we determine the reasons behind this result: countries with positive correlation are often visually similar to countries from different continents (e.g., $Tunisia$ is more similar to $Bolivia$ with an average similarity $0.94$ and distance $9,773$ than to $Austria$ with 
with an average similarity $0.80$ and distance $1,362$). We hypothesize this might be due to history, climate, and/or income differences, which could contribute more to visual similarity than distance alone. 
Our analysis shows that geographical location does not generally correlate with visual similarity. 
Therefore, collecting globally diverse annotations on a budget requires considering other complementary information, such as the country's income, culture, history, and climate. Our results on which countries are similar to each other provide valuable insights into how to distribute the annotation budget effectively and can be used along with this complementary information.

\paragraph{Augmenting with data from visually similar countries significantly improves model performance.}

We train a classifier to predict the topic of the input images and measure the accuracy while controlling for the countries.
Specifically, we input the CLIP visual representation in a linear layer, followed by a softmax to predict the topics of the input images.\footnote{We set the learning rate as 5e-3, use AdamW as the optimizer, and conduct training over 250 epochs with a batch size of 512. Additionally, we use a cosine annealing schedule with 50 warm-up epochs.}
We select one random country for each topic from the low-resource data, which we call target (topic, country) pairs. 
Next, we split the data into training and test sets in a 90-10\% data split to include all the target (topic, country) pairs in both sets. 
Finally, we replace different ratios (100\%, 90\%, 70\%, 50\%, 30\%, 10\%, 0\%) of the target-country data with images from: (1) the most \textit{similar} countries to the target-country given the target-topic; (2) the most \textit{dissimilar} countries to the target-country given the target-topic; (3) \textit{high-resource} data corresponding to the target-topic.

\begin{figure}[h!]
    \centering
    \includegraphics[width=0.9\textwidth]{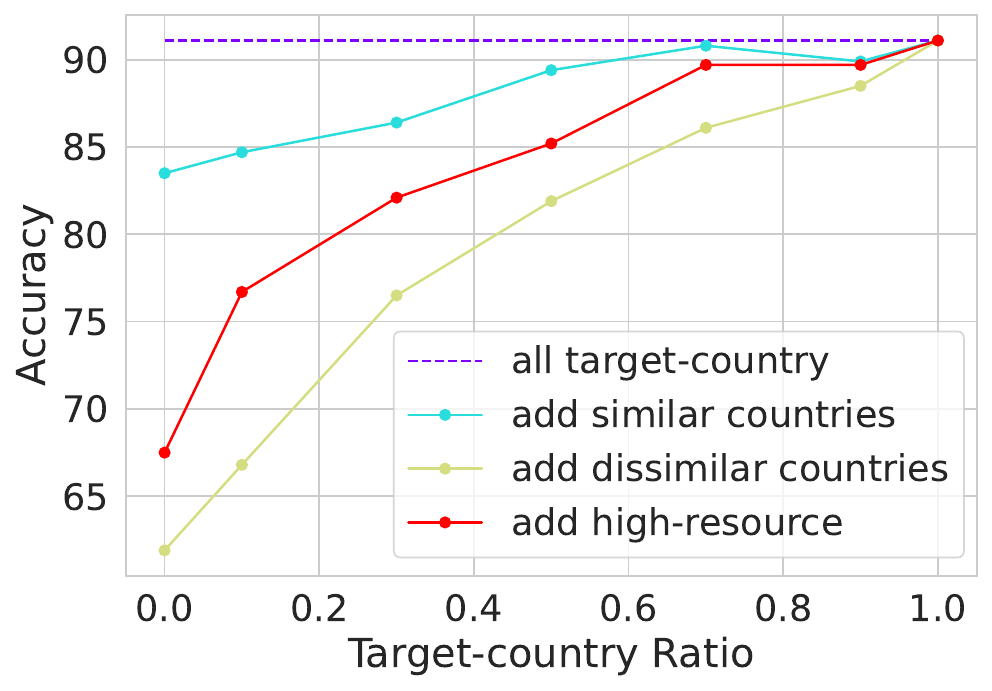}
    \caption{Topic classification accuracy (in \%) for different target-country data ratios (e.g., target-country ratio 0.0\% is equivalent to 100\% replacement ratio).
    We replace different ratios of the target-country data with images from: (1) the most similar countries to the target-country given the target-topic; (2) the most dissimilar countries to the target-country given the target-topic; (3) high-resource data of the target-topic;
}
    \label{fig:accuracy}
\end{figure}

The topic classification accuracy when using all the training target-country data is $91.1\%$, which is an upper bound. In  \Cref{fig:accuracy}, we show the accuracy when adding data from (1), (2) and (3).
The main takeaway is that \textbf{adding data from \textit{similar} countries improves the performance more than adding data from \textit{dissimilar} countries or \textit{high-resource} data, and the gap in performance increases with the replacement ratio}. Additionally, supplementing with \textit{high-resource} data is generally more beneficial than supplementing with data from \textit{dissimilar} countries.
We also compute the accuracy when no data is added, and find that adding data from \textit{dissimilar} countries or from \textit{high-resource} data can hurt the performance compared to not adding data, especially for high replacement ratios 
($50\%-90\%$). We show the results in the Appendix, in \Cref{fig:accuracy2}.

\section{Main Takeaways}
\begin{figure}[h]
    \centering
    \includegraphics[width=\textwidth]{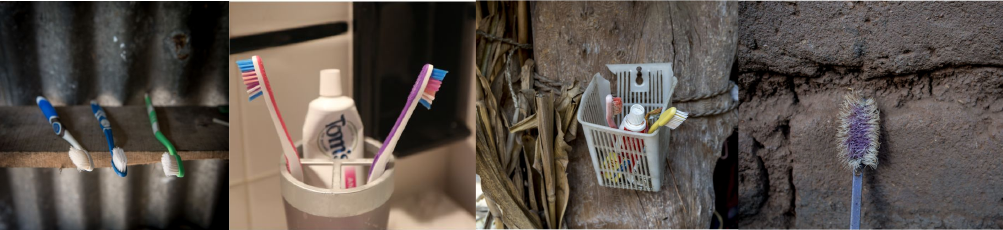}
    \caption{The context of the topic influences the visual similarity. For example, although the same type of toothbrush is depicted, their storage place differs, i.e., on a piece of wood, in a plastic container in the bathroom, in a plastic container tied to a tree, near a brick wall. Therefore, visual diversity is measured not only at the topic level but also at the context level.}
    \label{fig:visual_context_diff}
\end{figure}

Our analyses provide multiple insights into the current state of vision-language annotations for various topics across different countries, and show the coverage limitations of existing large-scale datasets. 
We highlight the main takeaways and propose actionable steps to help future work create more inclusive datasets and models.
\vspace{-4mm}
\paragraph{We recommend focusing the annotation efforts on currently underrepresented data.} To have more inclusive models and datasets, we need to collect more globally diverse annotations. Because annotations are expensive, we propose to focus future annotation efforts on specific countries and their topics.
To assist with these efforts, we provide a list of countries and corresponding topics that are consistently unrepresented in the training data of vision-language models. 
Furthermore, most countries have less than ten images per topic.
For most countries and corresponding topics -- 3,329/ 4,830, we could not determine how similar they are to the high-resource data because of the lack of data. These countries have less than ten images per topic and, therefore, already need annotations.
As an alternative solution, we recommend developing algorithms that can perform well with limited amount of data. 
\vspace{-4mm}
\paragraph{We can leverage cross-country data similarity to supplement data from unrepresented countries effectively.}
When we do not have a sufficient budget to annotate more data for a target country and topic, we propose using the available data from countries with similar visual representations of that given topic. We provide a list of similar countries for each target country and topic and show that using this data improves model performance more than using data from dissimilar countries or high-resource data.
\vspace{-4mm}
\paragraph{Geographical distance does not correlate with visual similarity between countries.}
We compute the Pearson correlation coefficient between the visual similarity and the geographical distance between all countries and find a very weak negative correlation of -0.01. Therefore, collecting globally diverse annotations requires considering additional information.
Multiple other factors, such as income, history, or cultural heritage, can contribute to the visual similarity between countries. We find this hypothesis worth investigating in depth in future work.
\vspace{-4mm}
\paragraph{Visual similarity between countries and topics depends on the context.}
While examining images of topics across countries, we notice visually similar topics with very different backgrounds, which influence the visual similarity score. For example, in ~\Cref{fig:visual_context_diff}, many countries have the same type of toothbrush, but because their storage place is different, their visual similarity score is low. In this paper, we measure similarity at the context level, considering both the topic and the context (e.g., background, storage space). However, as future work, we propose to investigate further which type of similarity to consider when we annotate diverse data: either at the topic level, by extracting the segmentation mask of the topic, or at the context level, by considering the entire image.
    
\section{Conclusion}
In this paper, we addressed the need for balanced data representation used to train vision-language models. Because data annotations are expensive, we proposed to annotate primarily images from unrepresented countries. To find which countries are less represented in the training data of vision-language models, we compared the visual similarity of images across 94 topics and 52 countries found in crowd-sourced and web-scraped data. We used three visual representations, CLIP, BLIP-2, and ALIGN, to ensure the results generalize across representation types.
Additionally, we proposed to leverage cross-country data similarity to improve model performance. We found visually similar countries for each country and corresponding topics and made them available in our repository:
{\UrlFont \url{https://github.com/MichiganNLP/visual_diversity_budget}}. 
Finally, our analysis offers multiple takeaways for future work to make informed decisions on what global data to annotate and how to leverage cross-country data similarity to improve model representation.
Through our work, we hope to contribute to building more inclusive and affordable vision-language models and datasets to help democratize AI globally.

    


\section*{Acknowledgments}
We thank the anonymous reviewers for their constructive feedback and are also grateful to the members of the Language and Information Technologies (LIT) lab at the University of Michigan for the insightful discussions during the project's early stages. This material is partly based on work supported by the Automotive Research Center (``ARC'') at the University of Michigan. Any opinions, findings, conclusions, or recommendations expressed in this material are those of the authors and do not necessarily reflect the views of ARC or any other related entity.

\nocite{*}
\section{Bibliographical References}\label{sec:reference}

\bibliographystyle{lrec-coling2024-natbib}
\bibliography{main.bib}

\appendix

\section{Subjective Topics} \label{sec:subj}
The 19 subjective topics that we remove: ``favorite home decorations'', ``favourite item in kitchen'', ``favourite sports clubs'', ``how the most loved item is used'', ``icons'', ``idols'', ``latest furniture bought'', ``looking over the shoulder'', ``most loved item'', ``most loved toy'', ``most played songs on the radio'', ``music idol'', ``next big thing you are planning to buy'', ``playing with most loved toy'', ``thing I dream about having'', ``things I wish I had'', ``using most loved item'', ``youth culture'', ``what I wish I could buy''.

\section{Data Stats}
\begin{figure*}[h!]
  \centering
  \includegraphics[width=0.98\textwidth]{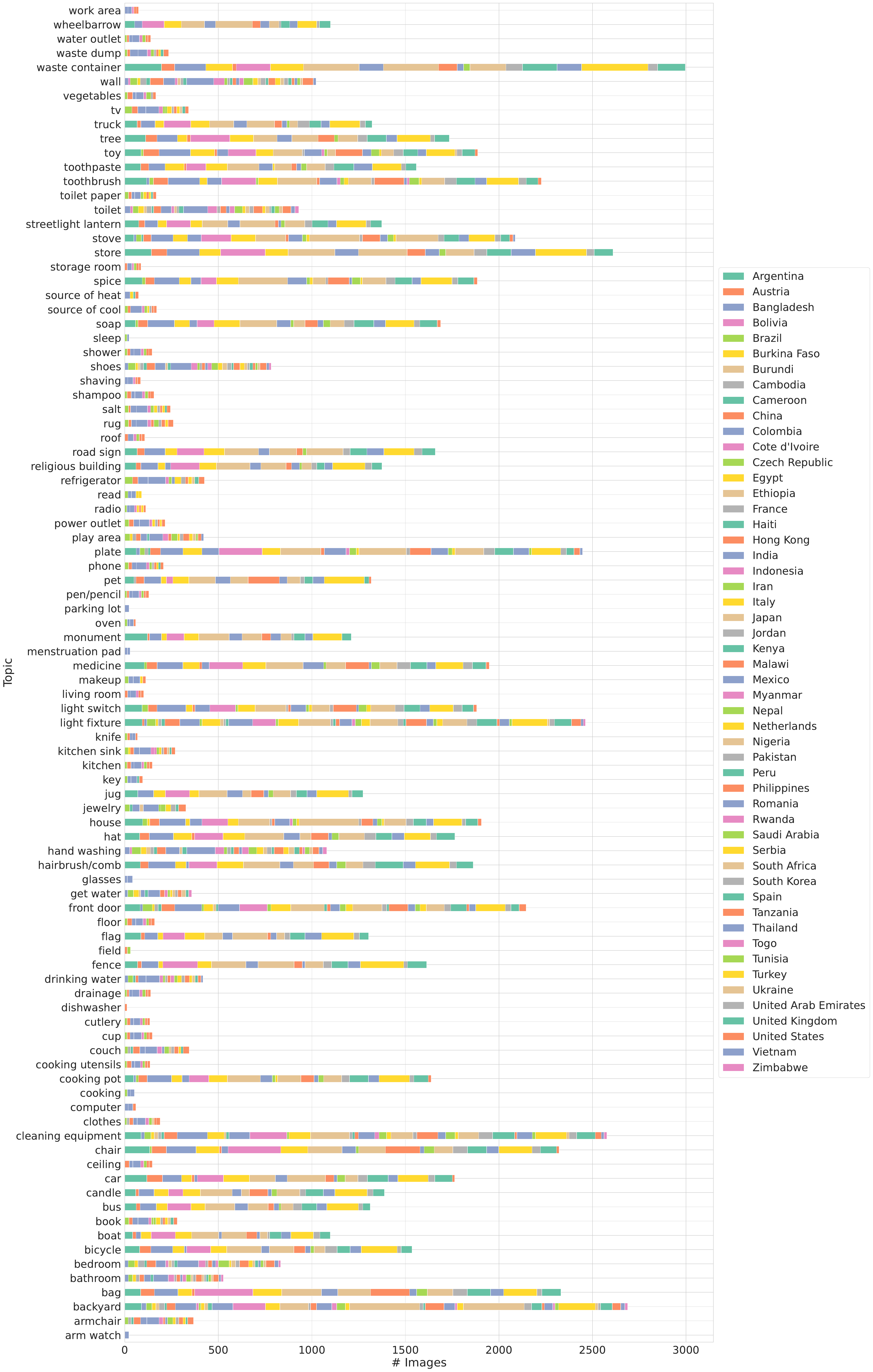}
  \caption{The distribution of countries per topic.}
  \label{fig:topic_distrib}
\end{figure*}

\clearpage

\section{Research Question 1}

\subsection{$USA$ respresentations.}
\begin{figure}[h!]
  \centering
  \includegraphics[width=\textwidth]{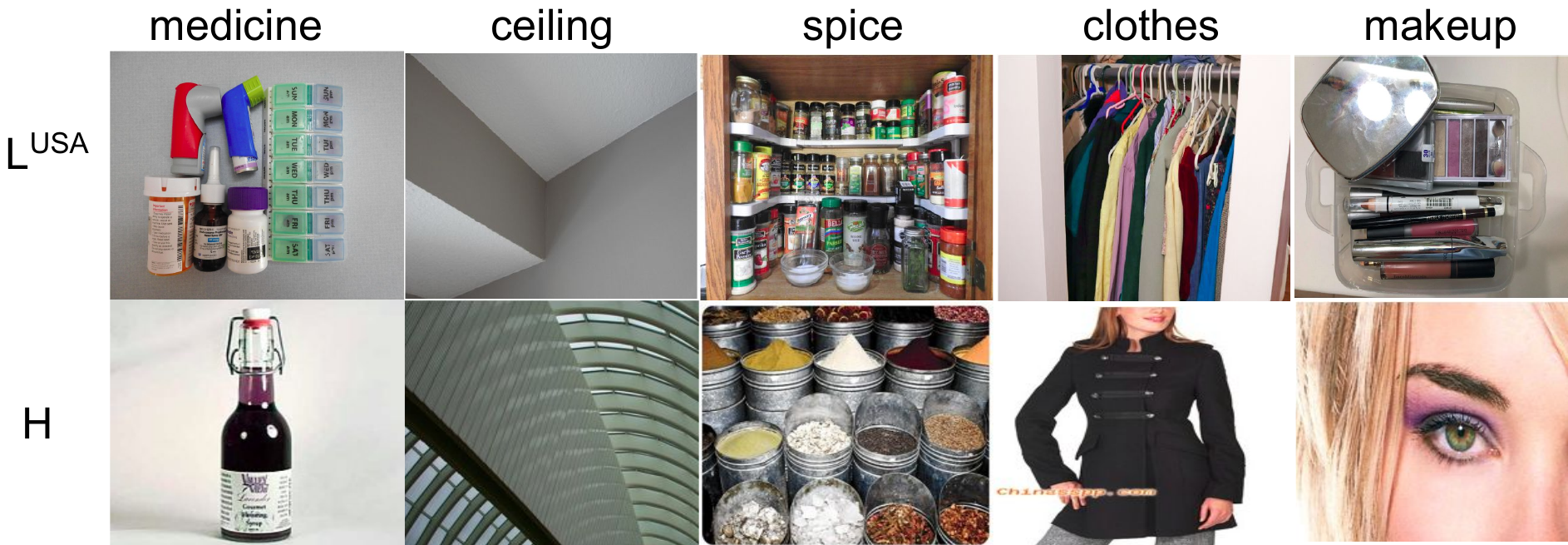}
  \caption{Representative images from the visually different topics in  low-resource $USA$ data/ $L$, and high-resource data/ $H$.
  In $H$, ``clothes'' and ``makeup'' are shown on people, while in $L$ they are separated in dressers and containers; in $H$, ``spice'' is in in large baskets in markets, while in $L$ they are in small containers in people's houses; in $H$, ``ceiling'' is shown in public spaces, while in $L$ is in private homes; in $H$ ``medicine'' is usually in bottles, while in $L$ can be in various forms.}
  \label{fig:usa_high_resource}
\end{figure}

\subsection{The effect of data size on the data analysis results.}

In \Cref{fig:subset_sim}, $Burundi$ has the lowest similarity score of 0.775 and has very little data in the heatmap of \Cref{fig:CLIP_avg_results_highlight}, only 97 images. 
Note however there are many counter-examples worth considering, such as countries with fewer images and high similarity scores (e.g., $Austria$ has eleven images and a similarity score of 0.862, $Bolivia$ has 37 images and a similarity score of 0.874), or on the opposing spectrum, countries with more images and low similarity scores (e.g., $Malawi$ has 390 images and a similarity of 0.776, $Burkina$ $Faso$ has 253 images and a similarity score of 0.789).

Furthermore, at topic level, the similarity scores and data size are not correlated (Pearson correlation score is -0.02). Similar to the country level, there are topics with many images and low similarity scores (e.g., $religious$ $building$ has 1,375 images and a similarity of 0.764) and topics with few images and high similarity scores (e.g., $glasses$ has 42 images and a similarity of 0.95).

In general, while data size can have an influence on our analysis results, we believe our work provides helpful strategies for annotation when data size is insufficient. Our paper is  a call to action for future work to collect more globally diverse data to improve the robustness of the results.

\begin{figure*}[h]
  \centering
  \includegraphics[width=\textwidth]{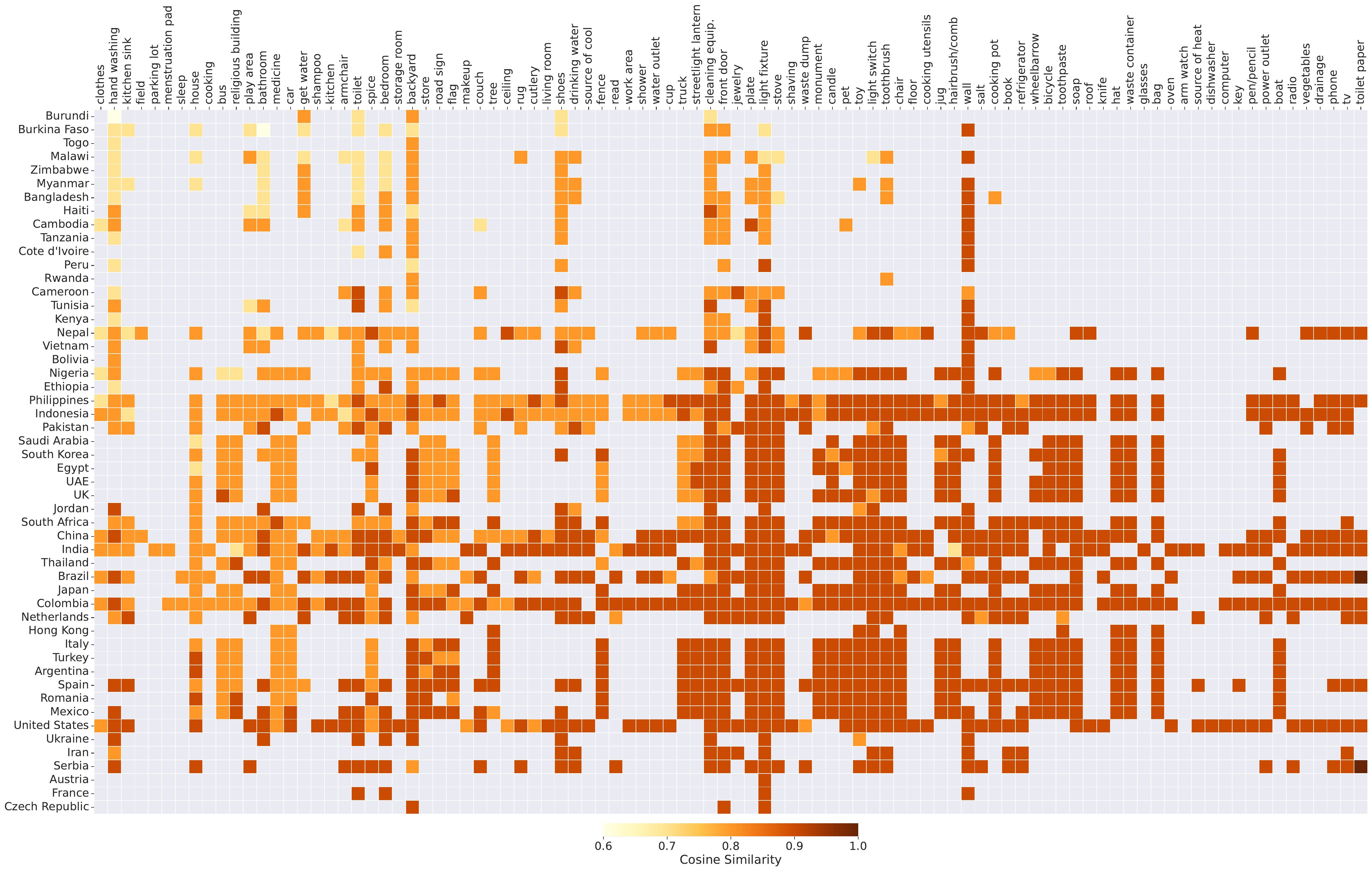}
  \caption{Similarity heatmap of (topic, country) pairs with CLIP visual representations. The darker, the less similarity between high-resource and low-resource data for that corresponding (topic, country), the more beneficial it is to annotate. Empty cells do not have any images for (topic, country). \textit{Best viewed in color.}}
  \label{fig:CLIP_avg_results}
\end{figure*}

\begin{figure*}[h!]
  \centering
  \includegraphics[width=\textwidth]{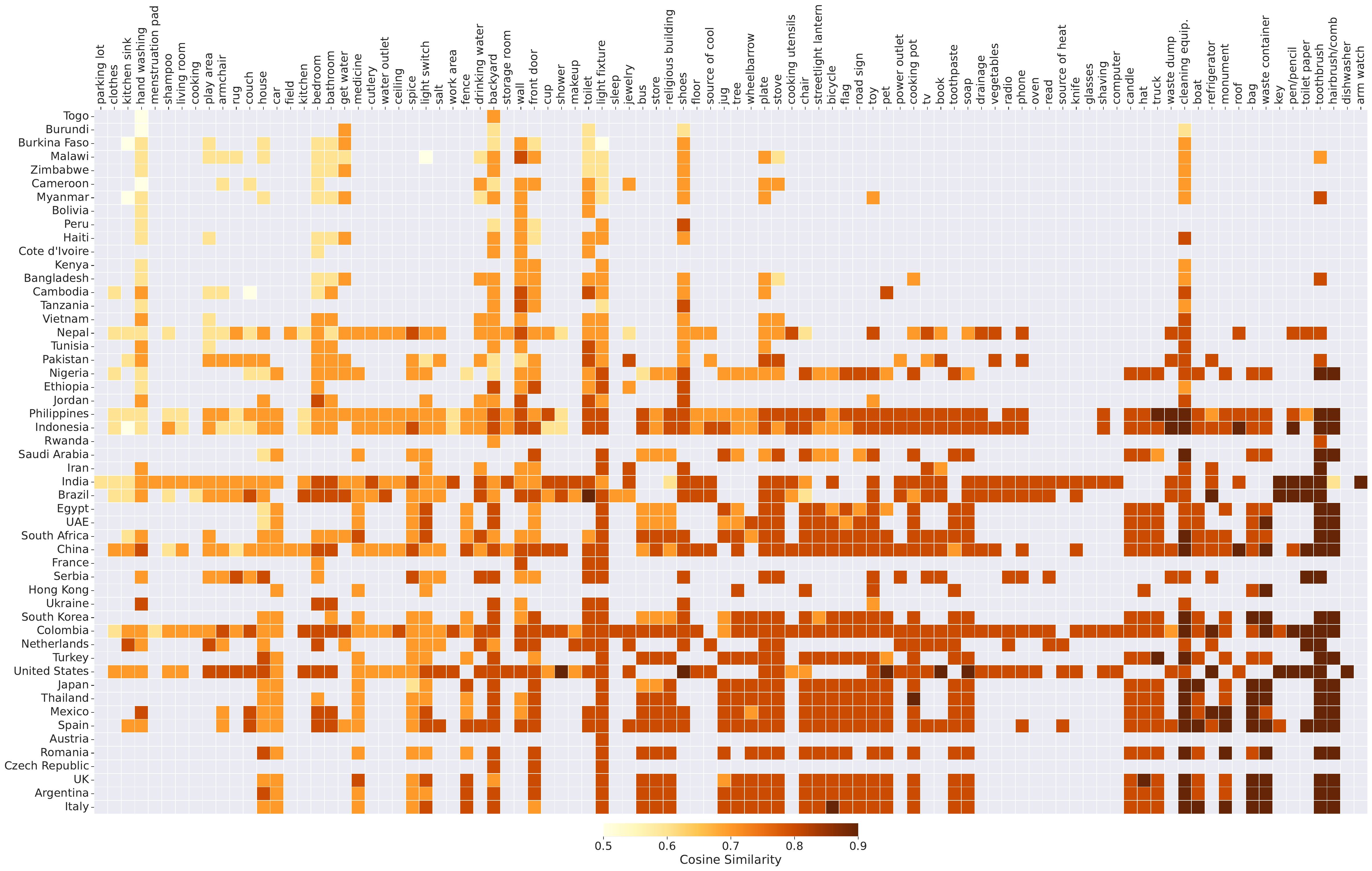}
  \caption{Similarity heatmap of (topic, country) pairs with ALIGN visual representations. The darker, the less similarity between high-resource and low-resource data for that corresponding (topic, country), the more beneficial it is to annotate. Empty cells do not have any images for (topic, country). \textit{Best viewed in color.}}
  \label{fig:ALIGN_avg_results}
\end{figure*}

\begin{figure*}[h!]
  \centering
  \includegraphics[width=\textwidth]{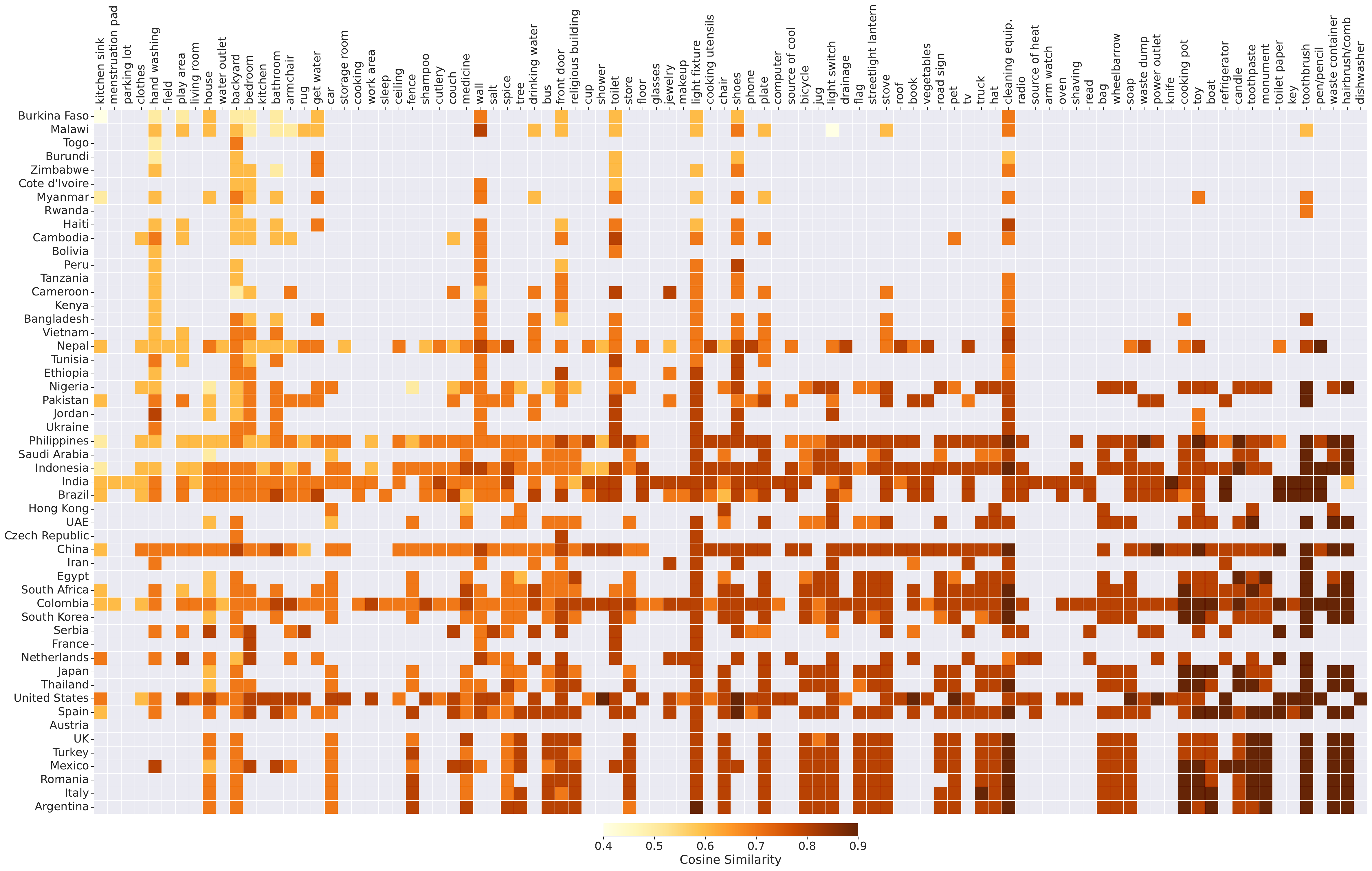}
  \caption{Similarity heatmap of (topic, country) pairs with BLIP visual representations. The darker, the less similarity between high-resource and low-resource data for that corresponding (topic, country), the more beneficial it is to annotate. Empty cells do not have any images for (topic, country). \textit{Best viewed in color.}}
  \label{fig:BLIP_avg_results}
\end{figure*}

\clearpage

\begin{figure}[h]
    \centering
    \includegraphics[width=\textwidth]{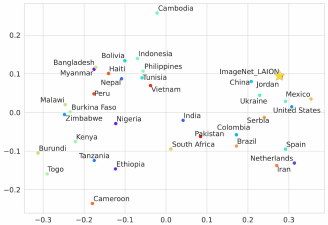}
    \caption{PCA for the topic ``hand washing'' for all countries that contain this topic in the low-resource data and in the high-resource data. The high-resource data point is highlighted. The data is represented as the average of the CLIP representations.}
    \label{fig:pca_hand-washing}
\end{figure}

\begin{figure}[h]
    \centering
    \includegraphics[width=\textwidth]{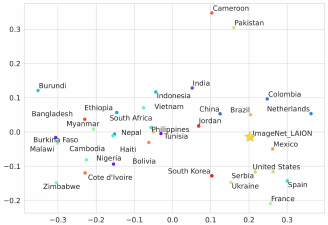}
    \caption{PCA for the topic ``toilet'' for all countries that contain this topic in the low-resource data and in the high-resource data. The high-resource data point is highlighted. The data is represented as the average of the CLIP representations.}
    \label{fig:pca_toilet}
\end{figure}

\begin{figure}[h]
    \centering
    \includegraphics[width=\textwidth]{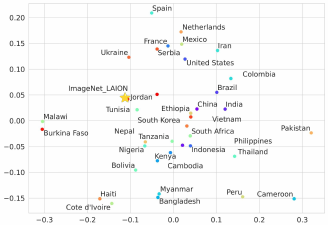}
    \caption{PCA for the topic ``wall'' for all countries that contain this topic in the low-resource data and in the high-resource data. The high-resource data point is highlighted. The data is represented as the average of the CLIP representations.}
    \label{fig:pca_wall}
\end{figure}

\clearpage

\section{Research Question 2}

\begin{figure}[h!]
  \centering
  \includegraphics[width=\textwidth]{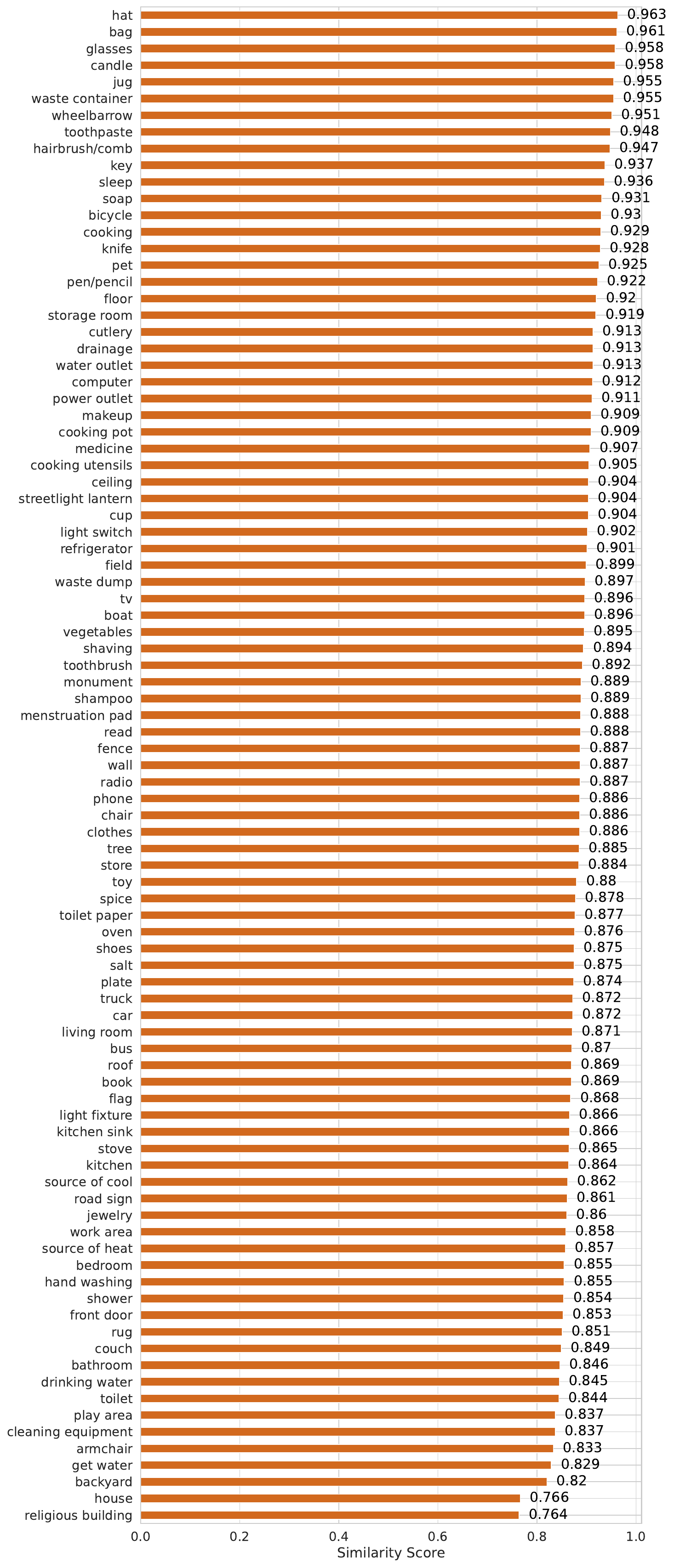}
  \caption{The distribution of average similarity scores per topic.}
  \label{fig:scores_topic}
\end{figure}

\begin{figure}[h!]
  \centering
  \includegraphics[width=\textwidth]{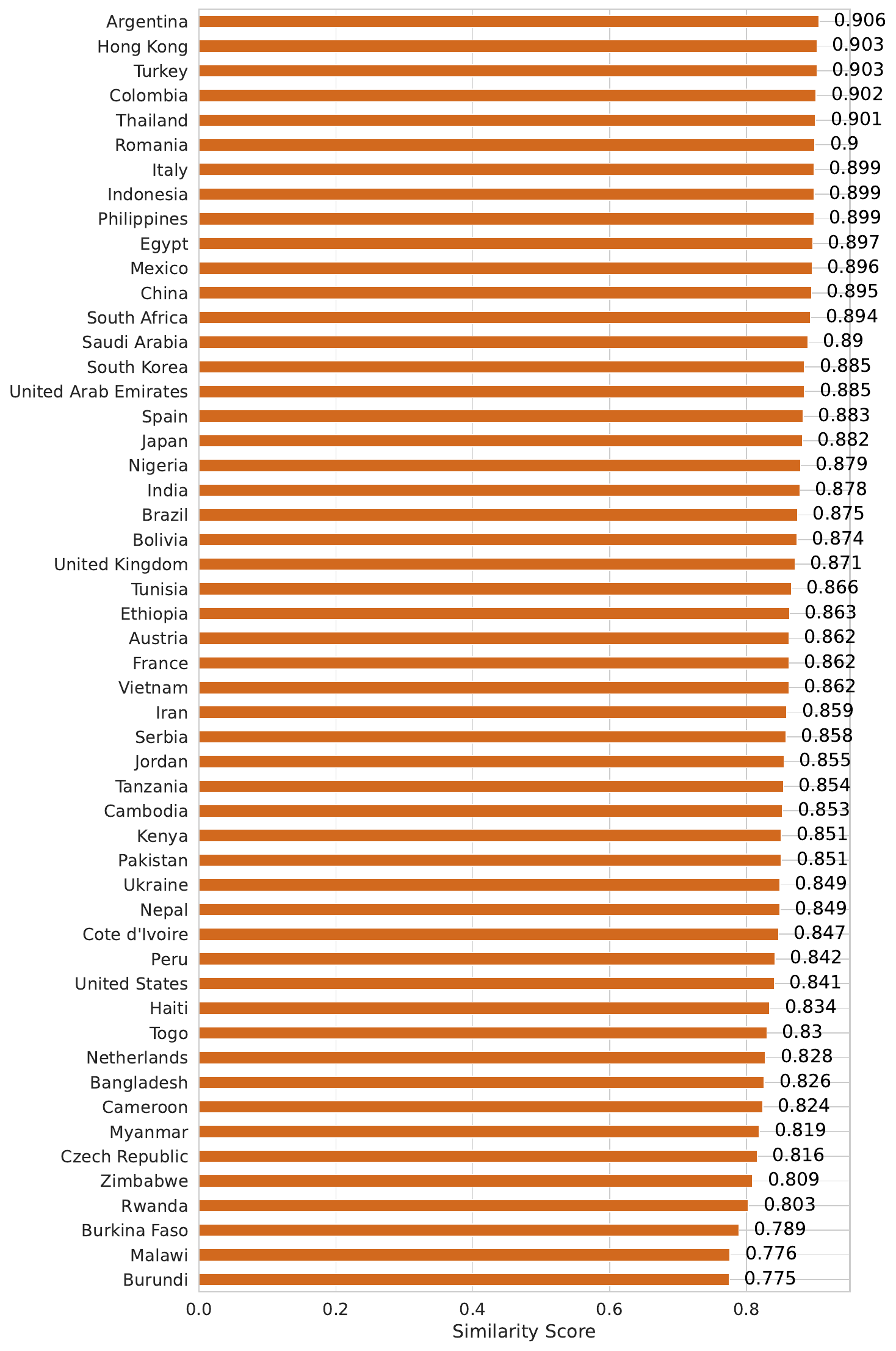}
  \caption{The distribution of average similarity scores per country.}
  \label{fig:scores_country}
\end{figure}

\clearpage

\begin{figure}[h!]
    \centering
    \includegraphics[width=\textwidth]{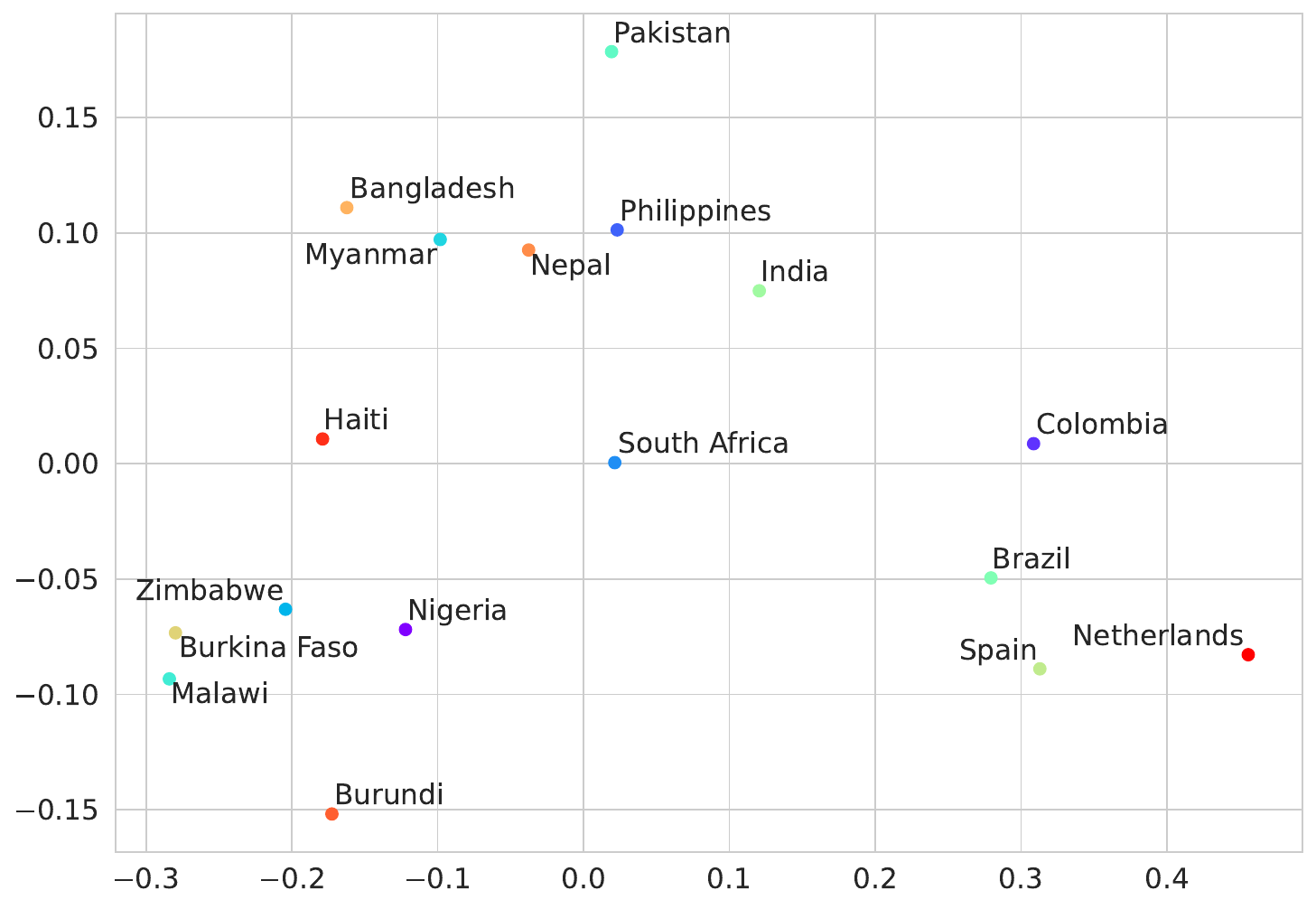}
    \caption{PCA for the topic ``get water'' for all countries that contain this topic in the low-resource data. The data is represented as the average of the CLIP representations.}
    \label{fig:get-water_PCA}
\end{figure}

\begin{figure}[h!]
    \centering
    \includegraphics[width=\textwidth]{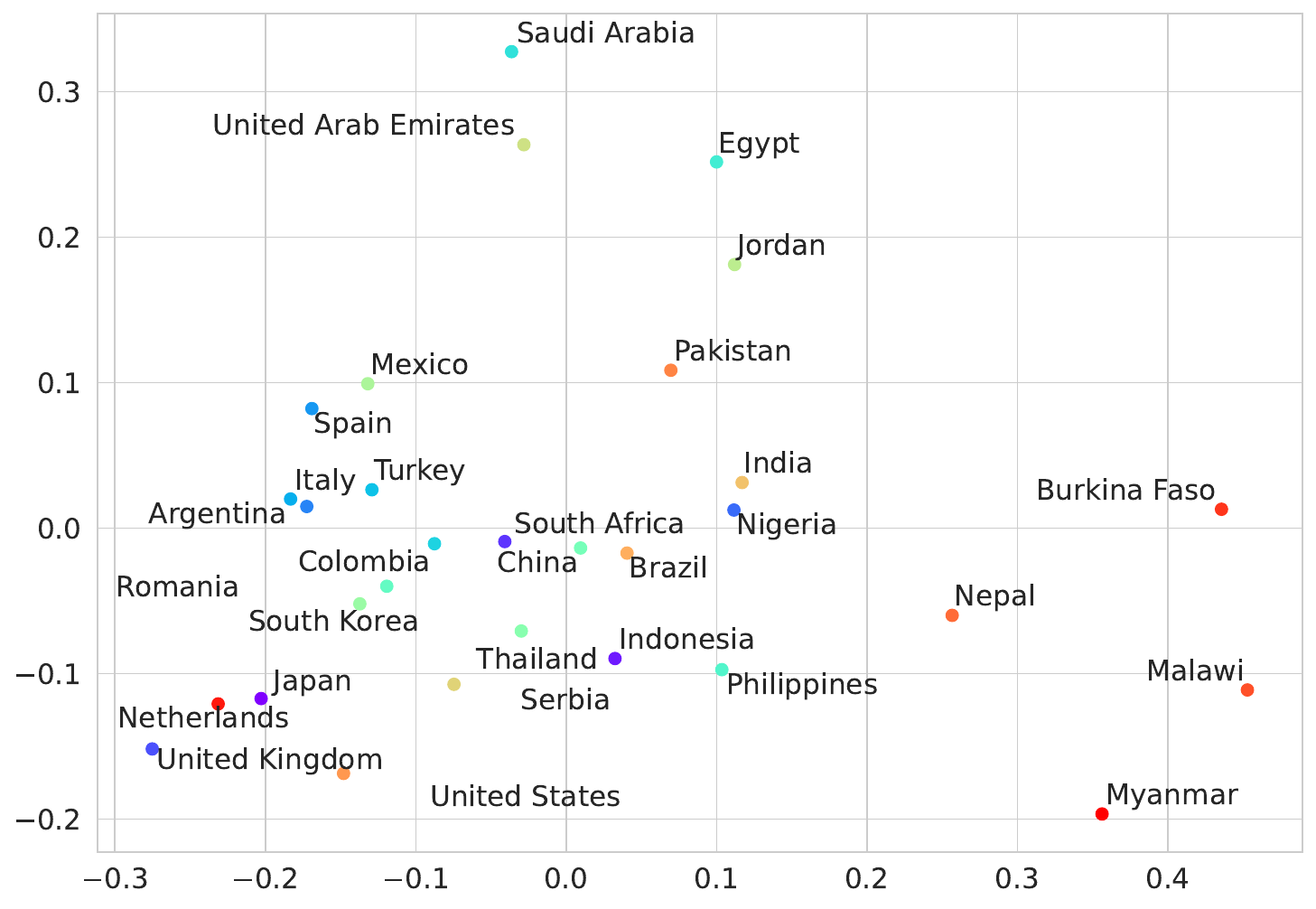}
    \caption{PCA for the topic ``house'' for all countries that contain this topic in the low-resource data. The data is represented as the average of the CLIP representations.}
    \label{fig:house_PCA}
\end{figure}

\begin{figure}[h!]
    \centering
    \includegraphics[width=\textwidth]{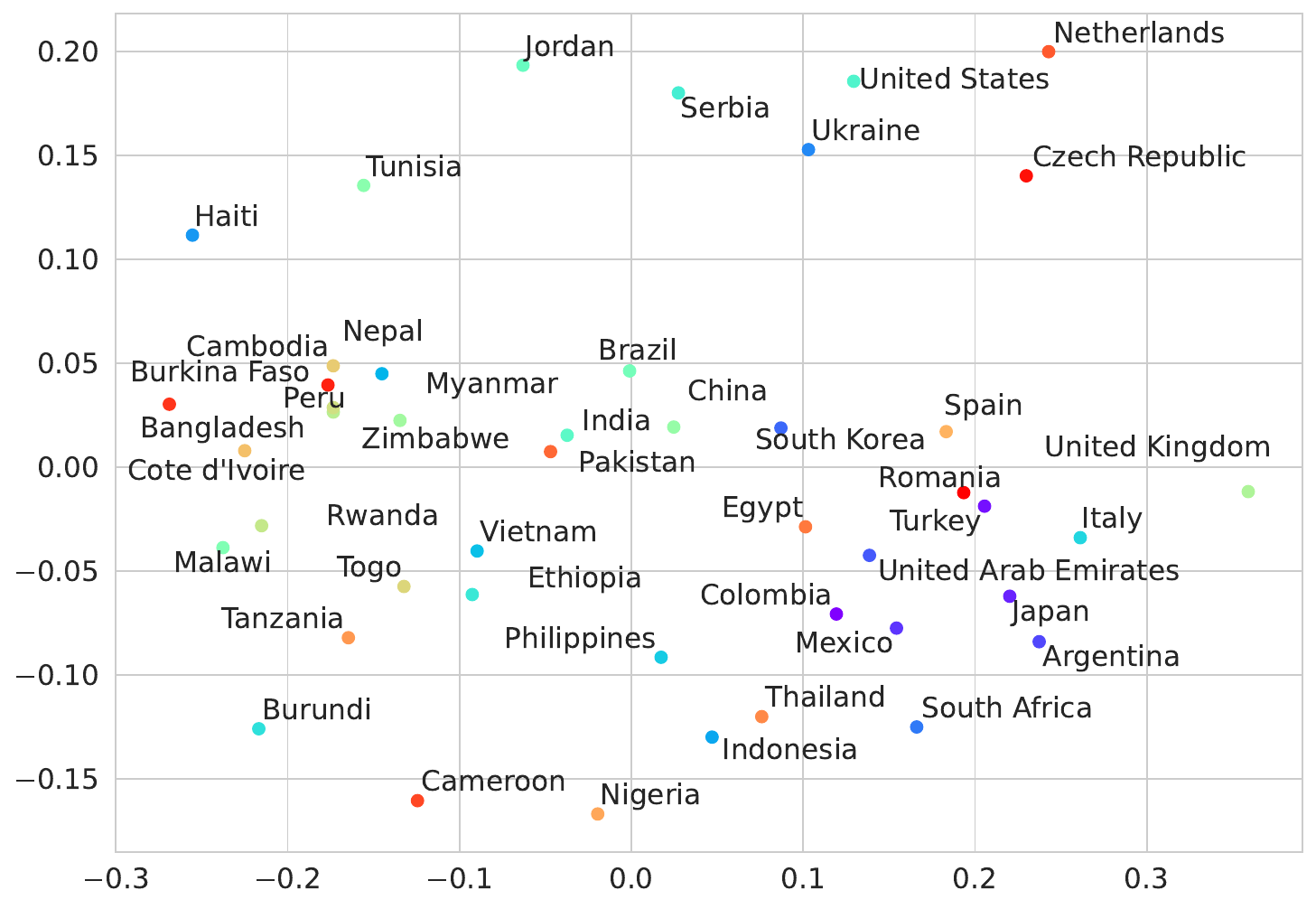}
    \caption{PCA for the topic ``backyard'' for all countries that contain this topic in the low-resource data. The data is represented as the average of the CLIP representations.}
    \label{fig:backyard_PCA}
\end{figure}


\begin{figure}[h!]
    \centering
    \includegraphics[width=0.8\textwidth]{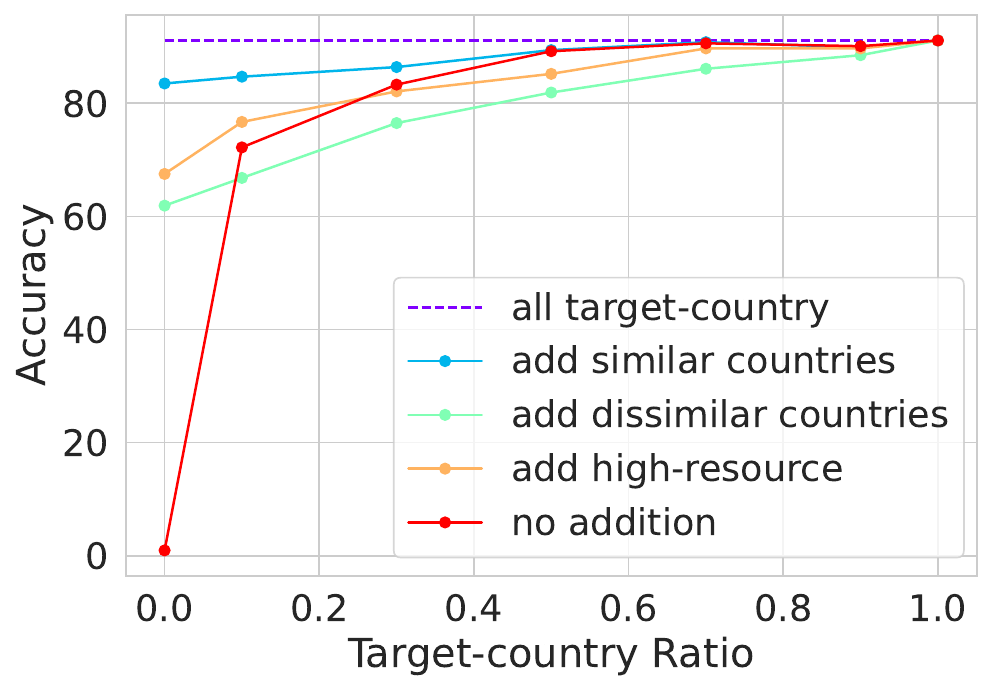}
    \caption{Topic classification accuracy (in \%) for different target-country data ratios (e.g., target-country ratio 0.0\% is equivalent to 100\% replacement ratio). We replace different ratios of the target-country data with images from: (1) the most similar countries to the target-country given the target-topic; (2) the most dissimilar countries to the target-country given the target-topic; 
(3) high-resource data of the target-topic;  (4) no replacement data/ addition.
}
    \label{fig:accuracy2}
\end{figure}

\end{document}